\documentclass[10pt,twocolumn,letterpaper]{article}

\usepackage[pagenumbers]{iccv}    

\definecolor{iccvblue}{rgb}{0.21,0.49,0.74}
\usepackage[pagebackref,breaklinks,colorlinks,allcolors=iccvblue]{hyperref}

\usepackage{graphicx}
\usepackage{amsmath}
\usepackage{amssymb}
\usepackage{booktabs}
\usepackage{tikz}
\usepackage{pgfplots}
\usepackage{multirow}
\usepackage{color}
\usepackage{array}
\usepackage[utf8]{inputenc}
\usepackage{times}
\usepackage{epsfig}
\usepackage{lipsum}
\usepackage{subfiles}
\usepackage{subcaption}
\pgfplotsset{compat=1.16}
\usepackage{enumitem}
\usepackage{float}
\usepackage{bm}
\usepackage{cuted}

\def\paperID{8208} 
\def\confName{ICCV}
\def\confYear{2025}

\title{Holistic Tokenizer for Autoregressive Image Generation}

\author{ 
Anlin Zheng\textsuperscript{\rm 1},
Haochen Wang\textsuperscript{\rm 2},
Yucheng Zhao\textsuperscript{\rm 3},
Weipeng Deng\textsuperscript{\rm 1}, \\
Tiancai Wang\textsuperscript{\rm 3},
Xiangyu Zhang\textsuperscript{\rm 4,5},
Xiaojuan Qi\textsuperscript{\rm 1}\thanks{Corresponding author: \texttt{xjqi@eee.hku.hk}} \\
\textsuperscript{\rm 1} The University of Hong Kong  \quad
\textsuperscript{\rm 2} NLPR, MAIS, CASIA \\
\textsuperscript{\rm 3} Dexmal \quad \textsuperscript{\rm 4} StepFun \quad
\textsuperscript{\rm 5} MEGVII Techonology \\
}

\begin{document}
\maketitle
\begin{abstract}


Vanilla autoregressive image generation models generate visual tokens step-by-step, limiting their ability to capture holistic relationships among token sequences. Moreover, because most visual tokenizers map local image patches into latent tokens, global information is limited. To address this, we introduce \textit{Hita}, a novel image tokenizer for autoregressive (AR) image generation. It introduces a holistic-to-local tokenization scheme with learnable holistic queries and local patch tokens. Hita incorporates two key strategies to better align with the AR generation process: 1) {arranging} a sequential structure with holistic tokens at the beginning, followed by patch-level tokens, and using causal attention to maintain awareness of previous tokens; and 2) adopting a lightweight fusion module before feeding the de-quantized tokens into the decoder to control information flow and prioritize holistic tokens. Extensive experiments show that Hita accelerates the training speed of AR generators and outperforms those trained with vanilla tokenizers, achieving \textbf{2.59 FID} and \textbf{281.9 IS} on the ImageNet benchmark. Detailed analysis of the holistic representation highlights its ability to capture global image properties, such as textures, materials, and shapes. Additionally, Hita also demonstrates effectiveness in zero-shot style transfer and image in-painting. The code is available at \href{https://github.com/CVMI-Lab/Hita}{https://github.com/CVMI-Lab/Hita}.

\end{abstract}    
\vspace{-1.5pc}
\section{Introduction}
\label{sec:intro}

Within an autoregressive (AR) generation paradigm, the field of large language models (LLMs) \cite{vaswani2017attention,bert,raffel2020exploring,radford2019language,brown2020language,zhang2022opt,wang2025ross} has witnessed significant progress in recent years. In particular, using AR generation along with transformer backbones, GPT-style models have exhibited impressive performance \cite{chatgpt,bard,bai2023qwen}, incredible scalability \cite{kaplan2020scaling,henighan2020scaling}, and versatile flexibility \cite{radford2019language,brown2020language} across various language tasks.

\begin{figure}[htpb]
  \centering
  \includegraphics[width=0.475\textwidth]{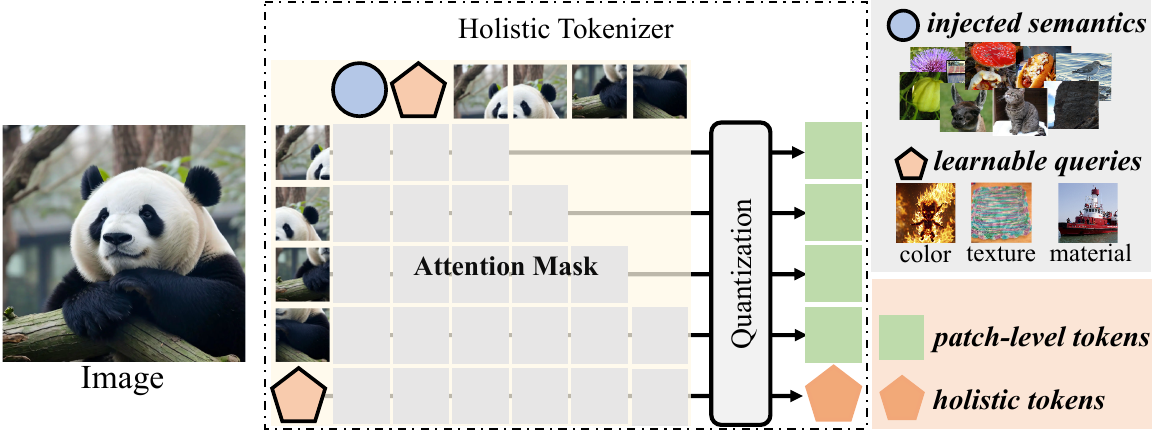}
  \vspace{-1pc}
\caption{\textbf{The concept of holistic tokenizer.} A set of learnable queries that capture global properties, such as color, texture, material, \textit{etc}, from pixels, with semantic-level feature injected, is utilized to reconstruct the image along with image patches.}
\label{fig:concept}
\vspace{-1.92pc}
\end{figure}

\begin{figure*}[t!]
  \centering
  \includegraphics[width=1.0\textwidth]{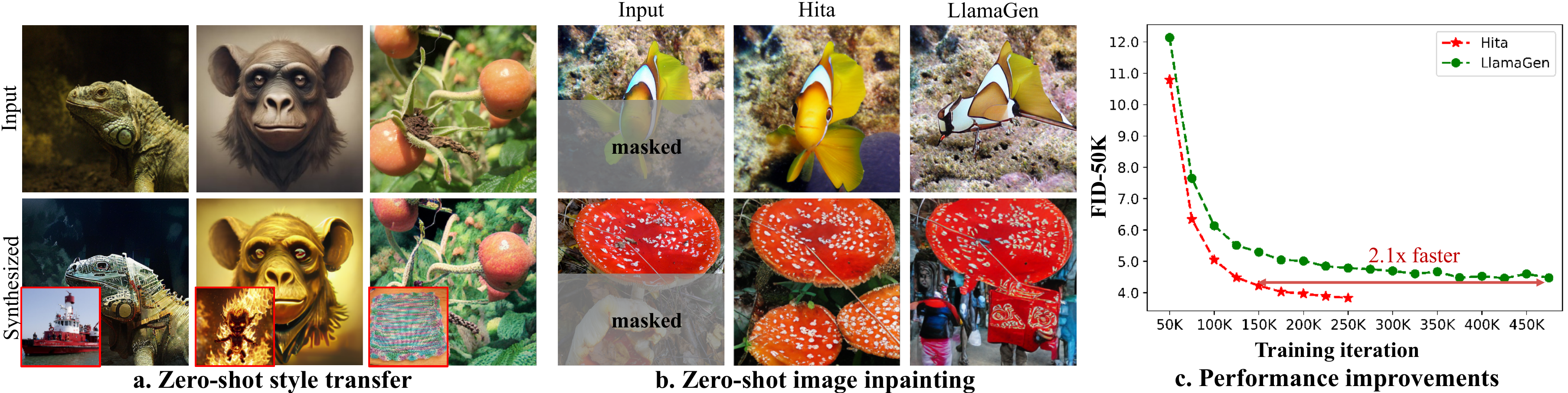}
  \vspace{-1.7pc}
\caption{\textbf{Holistic tokens} capture the global style and content information, as demonstrated by their ability to enable \textbf{(a)}  style transfer by replacing the holistic tokens of input images with those of reference images, and \textbf{(b)} zero-shot image inpainting using pre-trained AR generation models . Additionally, the introduction of these new latent tokens accelerates the training speed of the GPT-style AR image generation model by a factor of two \textbf{(c)}.}
\label{fig:primary-figure}
\vspace{-1.7pc}
\end{figure*}

The success of autogressive models in language modeling has driven substantial interest in their application to image generation, giving rise to representative works such as DALL-E \cite{ramesh2021zero}, Parti \cite{yu2021vector,yu2022scaling}, VAR \cite{var}, and LlamaGen \cite{llamagen}. In AR-based image generation, VQVAE \cite{vqgan} is employed to encode images into discrete tokens. These tokens are then raster-scanned into a 1D sequence to train an AR transformer model like Llama \cite{touvron2023llama}, which predicts the next token in the sequence. During inference stage, the token sequence generated by the AR model is decoded by the VQVAE's \cite{vqgan} decoder to produce the synthesized images.

While significant advancements have been made, most research efforts have concentrated on refining and scaling model architectures or optimizing learning objectives for either the VQVAE stage \cite{llamagen,yu2021vector,yu2022scaling,yu2023language} or the AR transformer stage \cite{llamagen,ramesh2021zero}. However, in terms of latent space design, existing approaches predominantly rely on patch-level image representations at either single scales \cite{yu2022scaling,llamagen} or multiple scales \cite{var}, which limits the ability of latent tokens to capture global and semantic-level information.
Although these spatially structured tokens are compatible with diffusion models \cite{song2020denoising, dhariwal2021diffusion, ho2022cascaded, sdxl, sd2.0, peebles2023scalable}, which predict all tokens in parallel, they introduce challenges for AR-based models. Autoregressive models typically employ causal attention to predict tokens sequentially \cite{chen2020generative, pixelgpt, var, llamagen}. The absence of holistic information makes it difficult for these models to maintain global coherence, leading to limitations in their capabilities and increasing learning difficulties. As shown in Fig.~\ref{fig:primary-figure}\textcolor{iccvblue}{b} (LlamaGen\cite{llamagen}) with image inpainting, when half of an image is input into the AR model as a prefix prompt, requiring it to generate the remaining part, the completed images do not align well with the original content. This challenge has also been highlighted in works such as \cite{scalablellm,mar,show-o}, where bi-directional attention mechanisms are introduced to mitigate the issue, albeit with non-trivial modifications to the AR generation process. 

To address these issues, we introduce Hita, a novel global-to-local image tokenizer for AR image generation. First, we propose using learnable queries to capture the holistic information from the patch-level image embeddings. Meanwhile, a pre-trained foundation model, such as DINOv2~\cite{dinov2reg}, is adopted to inject semantics into the learned queries to enhance their representation. Then, we quantize the learned queries and patch-level embeddings {using two separate codebooks}, resulting in holistic tokens and patch-level tokens. Next, we fuse the de-quantized holistic and patch-level tokens using a novel token fusion module to reconstruct the image. This design can better prioritizes the holistic token and improving both image reconstruction and generation quality.

Once Hita is trained, a vanilla autoregressive (AR) transformer model, \textit{e.g.} Llama \cite{touvron2023llama}, can be seamlessly incorporated to generate both holistic and image patch tokens sequentially, which can be decoded into image facilitated by Hita. During the AR generation, the AR generator first generates holistic tokens, which encapsulate global image information and serve as a prefix prompt. These holistic tokens then guide the subsequent generation of local image patch tokens, alleviating difficulties and maintaining global coherence for improved generation quality.

Our in-depth ablation experiments confirm that the learned holistic tokens effectively capture  global image features, such as shape, color, texture, \textit{etc}, from pixels (see Fig.~\ref{fig:primary-figure}\textcolor{iccvblue}{a}). This is demonstrated by effortlessly enabling style transfer by combining holistic tokens from reference images with patch-level tokens from the source image. Moreover, without task-specific training, the model can be seamlessly applied to image inpainting by encoding the given incomplete image into latent tokens, which serve as prefix sequence prompts to the AR model, requiring it to generate visual tokens for completing the image (see Fig.~\ref{fig:primary-figure}\textcolor{iccvblue}{b}), demonstrating the capability of holistic tokens in ensuring global consistent synthesis. 

To further validate its effectiveness, we benchmark Hita for AR image generation using the leading Llama-based \cite{touvron2023llama} AR transformer model. Extensive experiments show that Hita accelerates the convergence speed during training and significantly improves synthesis quality (see Fig.~\ref{fig:primary-figure}\textcolor{iccvblue}{c}). Specifically, the time required to reach an $\text{FID}$ of {4.22} is reduced by a factor of \textbf{2.1}. Furthermore, our final AR model with 2B parameters achieves an $\text{FID}$ of \textbf{2.59} on the ImageNet 256×256 class-conditional image generation benchmark. This outperforms the well-known AR generation model LlamaGen~\cite{llamagen} with 3B parameters. It also surpass the popular diffusion model LDM-4\cite{sd2.0} by~\textbf{0.9} FID and~\textbf{27.1} IS, respectively.

In summary, our contributions are:
\begin{itemize}
    \item We propose a novel global-to-local image tokenizer for AR image generation that allows integration of AR generation models without modification.
    \item The proposed image tokenizer emerges with new characteristics, such as zero-shot style {transfer} and zero-shot image in-painting.
    \item {Experiments} on class-conditional image generation demonstrate that it accelerates convergence during training and significantly improves synthesis quality.
\end{itemize}

\vspace{-1pc}
\section{Related Work}
\vspace{-0.5pc}

\noindent\textbf{Image Tokenization using Autoencoders.}
Image tokenization aims to transform images into compact latent tokens, reducing computation and redundancy for generative models \cite{magvit, magvit2, var, llamagen}. VQVAEs \cite{vqvae, vqvae2} integrate vector quantization into the VAE framework \cite{kingma2013auto} to transform continuous visual signals into discrete tokens, thus enabling autoregressive modeling like LLMs. Based on VQVAEs \cite{vqvae, vqvae2}, VQGAN \cite{vqgan} improves compression and reconstruction by adding an adversarial loss, while ViT-VQGAN \cite{vit-vqgan} combines transformers with VQGAN. Further advancements, like RQ-VAE \cite{residual-vq} and MoVQ \cite{movq}, explore multiple vector quantization stages for better latent embeddings. Recently, MAGVIT-v2 \cite{magvit2} and FSQ \cite{fsq} introduced quantization methods that do not rely on a lookup codebook. VQGAN-LC \cite{vqgan-lc} clusters features from a pre-trained CLIP \cite{clip} model, expanding the token set to over 10,000 with 99\% utilization. In contrast to methods that focus on learning patch -- level latent tokens-- often lacking global context and coherency, which can limit the effectiveness of AR models-- our work emphasizes the importance of holistic representation in the latent tokens.

TiTok~\cite{titok} and VAR~\cite{var} are related efforts but differ from Hita in several aspects. TiTok creates compact 1D tokens with queries to reduce redundancy, while Hita uses queries to capture holistic properties for 1D holistic tokens and fuses them with 2D image patch tokens for image reconstruction. TiTok refines queries for 1D tokens tailored for reconstruction, whereas Hita extracts information from a pre-trained foundation model and patch features to capture global information. TiTok requires a complex training strategy, while Hita allows efficient single-stage training. Hita outperforms TiTok in AR generation and offers new features like zero-shot style transfer and inpainting. VAR~\cite{var} predicts global and local tokens indiscriminately, while Hita separates holistic and patch-level tokens, enriching holistic tokens with pre-trained foundation to form global representations and uses causal attention to align with the causal nature of AR models, which VAR lacks. Removing initial coarse-scale tokens seldom affects VAR's reconstruction, indicating those tokens are not crucial.

\noindent\textbf{Autoregressive Image Generation.} 
Recent advancements in GPT-style transformers have sparked significant interest in AR generation, leading to various representative approaches \cite{chen2020generative, pixelgpt, llamagen, residual-vq, var, liu2024can}. Like GPT \cite{gpt}, these models predict sequential visual tokens autoregressively. Early AR models \cite{chen2020generative, pixelgpt} generated sequences directly in pixel space, while current models \cite{residual-vq, vit-vqgan, var, llamagen} use tokenizers \cite{vqvae, vqvae2, vit-vqgan, vqgan} to compress images into discrete tokens, followed by causal transformers for next-token prediction and image reconstruction through the tokenizer’s decoder. Most tokenization techniques generate discrete visual tokens in a patch-wise fashion, assuming a 2D grid structure for image signals. While logical for spatially structured inputs, this patch-based approach limits the tokenizer’s ability to capture holistic information. This limitation is amplified in AR models, where causal attention’s step-by-step prediction struggles to capture long-range dependencies. Studies \cite{paligemma, scalablellm, mar} show that causal attention in image token generation underperforms compared to bidirectional attention.

To address this, recent works \cite{var, mar, show-o} have introduced bidirectional attention to AR models with promising results. VAR \cite{var} introduced next-scale prediction, enabling simultaneous token generation and allowing bidirectional attention within each scale. MAR \cite{mar} incorporated a BERT-style framework to integrate bidirectional attention, while Show-o \cite{show-o} combined unidirectional and bidirectional attention in a single transformer for both comprehension and generation. However, these approaches complicate the design of a universal transformer to unify multi-modal understanding and generation while maintaining the next-token prediction paradigm. In this work, we introduce a new global-to-local tokenization method and investigate how far we can advance AR generation while preserving the inherent design of language models. 

\vspace{-0.4pc}
\section{Method}
\label{sec:approach}
In this section, we first introduce the preliminaries related to image tokenizers and AR generation models. Then, we present the design of our holistic tokenizer step-by-step. Next, we discuss their characteristics.
\vspace{-0.6pc}
\begin{figure*}[htbp]
  \centering
  \includegraphics[width=1\textwidth]{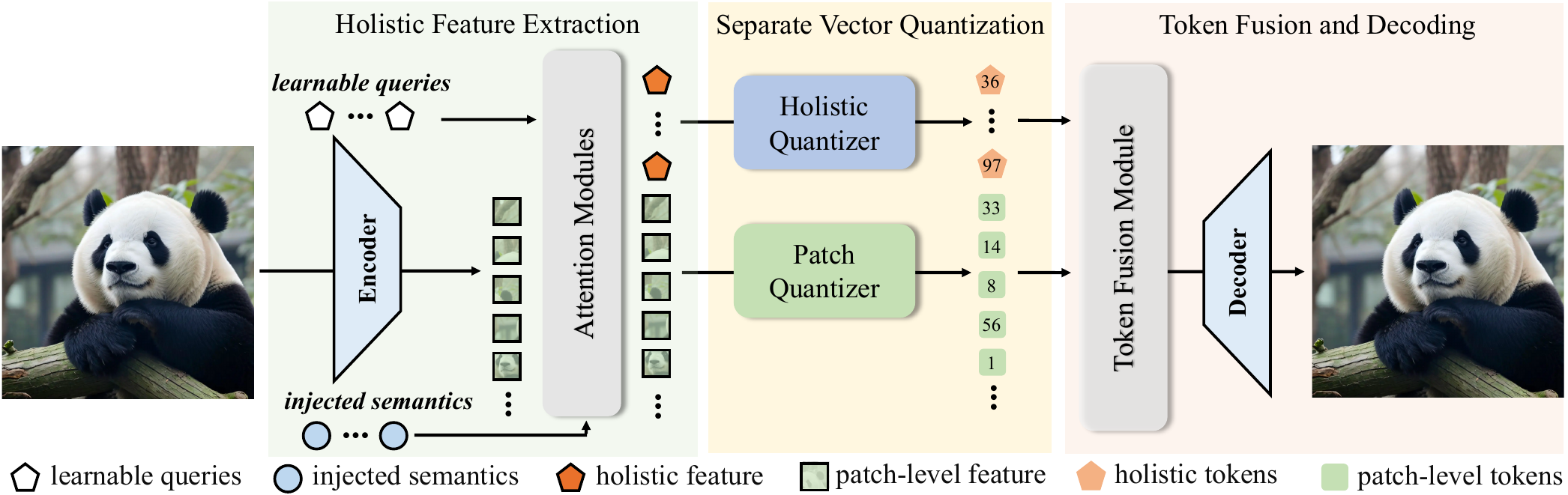}
\vspace{-1.5pc}
\caption{The flowchart of the proposed holistic image tokenizer, which includes:  \textit{Holistic feature extraction} part incorporates learnable queries to obtain holistic and patch-level features from pixels.  \textit{Separate vector quantization} part process holistic and patch tokens by separate codebooks. \textit{Token fusion and decoding} part leverages a lightweight token fusion module to fuse holistic tokens into patch-level tokens to reconstruct image. Here we omit de-quantization between quantizer and token fusion module for simplicity. }
\label{fig:framwork}
\vspace{-1.5pc}
\end{figure*}

\vspace{0.05in}
\subsection{Background}
\label{method:preliminary}

\vspace{-0.5pc}
\vspace{0.05in}\noindent\textbf{Quantized Image Tokenizer.}\label{vq-procedure}
To apply autoregressive modeling to visual generation, existing methods~\cite{ yu2021vector,yu2022scaling, llamagen, var} necessitate an image tokenizer to convert a 2D image into a 1D token sequences. Quantized autoencoders, such as VQVAEs \cite{vqvae, vqvae2,vqgan, vit-vqgan,vqgan-lc,var,llamagen}, are widely used. An image tokenizer generally consists of an encoder $\mathcal{E}(\cdot)$, a quantizer $\mathcal{VQ}(\cdot)$, and a decoder $\mathcal{D}(\cdot)$.  Given an input image $\text{I} \in \mathbb{R}^{H \times W \times 3}$, the encoder $\mathcal{E}(\cdot)$ first projects image pixels to the feature map $Z_{2D} \in \mathbb{R}^{\frac{H}{f} \times \frac{W}{f} \times \text{D}}$ with spatial down-sampling factor $f$. Then, $Z_{2D}$ is fed into the quantizer $\mathcal{VQ}(\cdot)$ that typically includes a learnable codebook $\mathbb{C} \in \mathbb{R}^{N \times D}$ with $N$ vectors. Each feature vector ${z_i} \in \mathbb{R}^{D}$ is mapped into its nearest vector ${c_i} \in \mathbb{R}^{D}$ in the codebook $\mathbb{C}$. This process can be formulated as Eq.~\eqref{equ:vq}.

\vspace{-0.13in}{\small \begin{equation}
\begin{split}
{Z_{2D}} &= \mathcal{E}(\text{I}), \\
\mathcal{VQ}({z_i}) = {c_i}, \quad \text{where} \quad {i} &= \mathop{\arg\min}\limits_{{j} \in \{1, 2, ..., N\}} \Vert {z_i} - {c_j} \Vert_2.
\end{split}\label{equ:vq}
\end{equation}}
where $H$ and $W$ denote the input image’s height and width respectively. $D$ depicts the latent feature dimension.  Once discrete tokens are acquired, they can be de-quantized into corresponding code and converted back to image pixels by the decoder $\mathcal{D}(\cdot)$, as depicted in Eq.~\eqref{equ:dec}.

\vspace{-0.13in}{ 
\begin{equation}
\label{equ:dec}
\mathbf{\hat{I}} = \mathcal{D}(\mathcal{VQ}({Z_{2D}})).
\end{equation}}

For the optimization of codebook, the training objective is $\mathbf{\mathcal{L}_{vq}}=\sum{\|\mathbf{sg}(z_i) -{c_i}\|^{2}_{2} + {\beta} \cdot \| \mathbf{sg}({c_i}) - {z_i}\|^{2}_{2}} $, where $\mathbf{sg}(\cdot)$ is a stop-gradient function~\cite{straightsthrough,vqvae}. The second term is a commitment loss with loss weight $\beta$ used to align extracted features with codebook vectors. For image reconstruction optimization, the loss function is $\mathcal{L}_{AE} = \mathcal{L}_2(\text{I}, \hat{\text{I}}) + \mathcal{L}_{P}(\text{I}, \hat{\text{I}}) + \lambda_{G} \cdot \mathcal{L}_{G}(\hat{\text{I}})$, where $\mathcal{L}_2$ is a pixel-wise reconstruction loss, $\mathcal{L}_{P}$ is perceptual loss from LPIPS \cite{percetualloss}, $\mathcal{L}_{G}$ is adversarial loss from PatchGAN \cite{patchgan} with weight $\lambda_{G}$.

\vspace{0.05in}\noindent\textbf{Autoregressive Image Generation.} \label{autoregressive models} 
When used for visual generation, {the 2D quantized map from a tokenizer is converted into a 1D sequence in a pre-defined order, \textit{e.g.} raster-scan order. Then, an AR generation model autoregressively predicts the sequence of image tokens. To achieve this, AR models typically incorporate causal attention, using a prefix token sequence to predict the next, focusing on learning the dependencies between the prefix and current discrete image tokens. Considering a sequence of discrete tokens ${x_{1:N}}$, an AR model is trained on these tokens by maximizing the cross-entropy depicted in Eq.~\eqref{equ:autoregressive}, where ${c}$ is class label or text embedding, ${x_{j<i}}$ serves as the prefix sequence of tokens for the current token ${x_i}$.
}
\vspace{-.1in}{\begin{equation}
\label{equ:autoregressive}
 \mathop{\max} \mathcal{L}^{\text{CE}}_{\theta} = \sum\limits_{i=1}^{N} \text{log} P_{\theta}({x_i}| {x_{j<i}}, c) 
\end{equation}}
During inference, the AR model initially predicts visual tokens autoregressively by sampling tokens from the learned distribution. Subsequently, the generated visual tokens are converted back into images through the decoder $\mathcal{D}(\cdot)$.
\subsection{Holistic Tokenization for AR Generation}
\label{method:hita}
As shown in Fig.~\ref{fig:framwork}, Hita incorporates the following key components. First, to capture holistic properties, we introduce learnable queries that interact with image patch embeddings via attention modules. A pre-trained foundation model (\textit{e.g.}, DINOv2 \cite{dinov2reg}) injects semantic-level features. To align Hita's latent space with the causal nature of AR models, we use a causal transformer to integrate holistic and patch features before quantization, embedding the sequential structure into the tokenization process for learning prefix-aware latent tokens.  Details are depicted in Sec.~\ref{sec:holisct_capture}. Then, the holistic and patch-level features are quantized using separate codebooks (see Sec.~\ref{sec:quantization}). Finally, after dequantization and during token fusion and decoding procedure, Hita first uses a fusion module with causal transformers to process these tokens in order and prioritize holistic tokens. Then they are fed into the decoder for image reconstruction (see Sec.~\ref{sec:shift_picking}). 

\subsubsection{Holistic Feature Extraction}
\label{sec:holisct_capture} 
We build Hita based on VQGAN \cite{vqgan}. Specifically, we adopt its encoder $\mathcal{E}(\cdot)$ to encode input image $I$ into local patch embeddings, and further incorporate a set of learnable queries ${Q}$ to gather holistic information from the whole input. To enhance the holistic representation, a pre-trained foundation model $\mathcal{H}(\cdot)$ like DINOv2 \cite{dinov2reg} is leveraged to inject rich semantic-level features.  To achieve this, we first concatenate the learnable queries and the flattened patch-level embeddings as well as foundation model features into sequence and feed them into a simple transformer $\mathcal{E}_{\text{{trans}}}(\cdot)$ is utilized, where holistic queries ${Q}$ interact with all patch-level image embeddings and the injected semantic-level features.  Since the AR model models the probability of observing the current token based on its predecessors, a latent space well-aligned with the causal nature is encouraged. Therefore, the extracted holistic and patch-level features are concatenated and reorganized in an ordered sequence, with holistic queries come first followed by the patch-level features obtained in raster-scan order. This sequence is then fed into a transformer with global features serving as prefix prior. The whole process is formulated as Eq. \eqref{equ:context}.

\vspace{-.3cm}
\begin{equation}
\begin{split}
     {Q},\ {Z} &= \mathcal{E}_{\text{trans}}({Q} \oplus \mathcal{E}({I}) \oplus \mathcal{H}({I})) \\
     {\overline{Q}},\ {\overline{Z}} &= \mathcal{E}_{\text{casual}}({Q} \oplus {Z})
\end{split}\label{equ:context}
\end{equation}
{where ${Z} \in \mathbb{R}^{\frac{H}{f} \times \frac{W}{f} \times {C}} $ refers to the path-level embeddings with a spatial downsample ratio $f$. ${Q} \in \mathbb{R}^{M \times C} $ denotes learnable queries. $\oplus$ is the concatenation operation. Moreover, the pre-trained feature is discarded after $\mathcal{E}_\text{trans}$.}

\begin{figure}[thpb]
  \centering
  \includegraphics[width=0.45\textwidth]{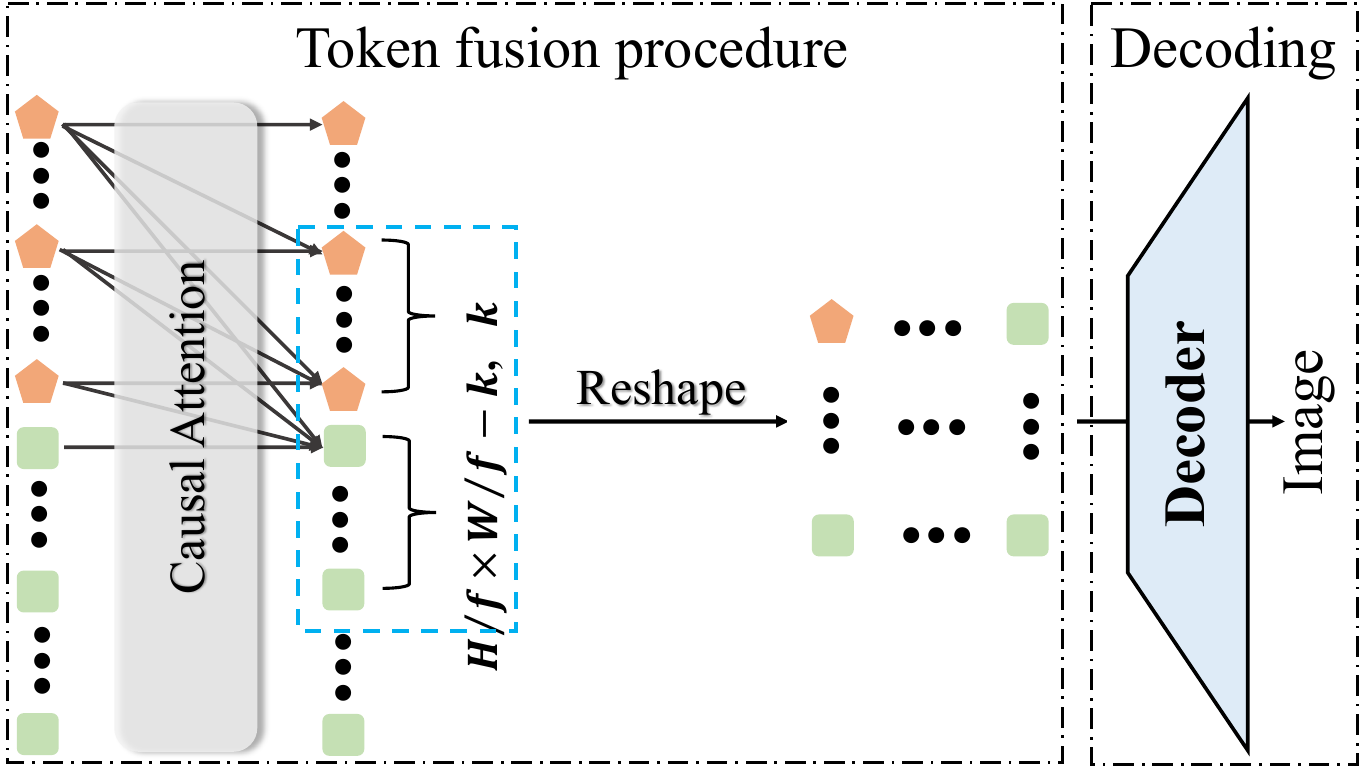}
  \vspace{-0.7pc}
\caption{Token fusion and decoding procedure}
\label{fig:token-fusion}
\vspace{-1.4pc}
\end{figure}
%

\subsubsection{Separate Vector Quantization}
\label{sec:quantization}

Once both learned queries $\overline{Q}$ and patch-level images $\overline{Z}$ are produced, they are quantized with holistic quantizer $\mathcal{Q}_H(\cdot)$ and patch quantizer $\mathcal{Q}_P(\cdot)$, respectively. Then, the holistic tokens $\hat{Q}$ and patch-level $\hat{Z}$ tokens could be obtained. The the codebook significantly impacts the performance of the image tokenizer. Following \cite{vit-vqgan, llamagen}, we apply ${\ell}_2$-normalization to the codebook vectors, opting for a lower vector dimension and a larger codebook size to improve the reconstruction quality and codebook utilization.

\subsubsection{Token Fusion and Decoding Procedure}
\label{sec:shift_picking}
After quantization, the holistic tokens and patch-level tokens can be fed into a standard transformer for token fusion and then fed into the decoder to reconstruct the image. However, such a design incurs codebook collapse in the holistic quantizer ({3th row in Table.~\ref{tbl:ablation_k}}). This occurs because patch-level tokens directly impact the reconstruction of relevant image patches through the skip connection in the transformer, which makes it bypass holistic tokens (see Fig.~\textcolor{iccvblue}{1a} in Appendix). As a result, the token fusion module learns a trivial solution that overlooks the holistic tokens and leads to holistic codebook collapse.

Inspired by the next-token prediction used in AR models, we first concatenate the holistic tokens and the patch-level tokens in an ordered sequence, placing the holistic tokens first, followed by the patch-level tokens flattened in a raster-scan order. Next, they are fed into an attention transformer for token fusion. After this, we choose the last $k$ holistic tokens, combine with the first $\frac{H}{f} \times \frac{W}{f} - k$ patch tokens to construct a 1D sequence and reshaped into a 2D grid by reshaping operation $\mathcal{R}(\cdot)$. It would be decoded back into an image $\hat{I}$ facilitated by the decoder $\mathcal{D(\cdot)}$. 
The motivation for this design is that, through this revision, the information carried by patch-level tokens flowing through skip connections becomes incomplete, requiring interaction with holistic tokens to compensate. This ensures that the token fusion process avoids trivial solutions and prevents the holistic codebook collapse, while emphasizing the importance of holistic tokens. Further discussion is also presented in the Appendix. Moreover, to better align the token sequence with the causal nature of the AR generation model, we adopt causal attention $\mathcal{\hat{E}}_\text{causal}(\cdot)$ to construct the token fusion module.
The whole procedure can also be formulated as Eq.~\eqref{equ:decoding} and illustrated in Fig.~\ref{fig:token-fusion}. Additionally, this design could better improve the reconstruction and generation quality.

\begin{equation}
\begin{split}
        {\tilde{Q}},\ {\tilde{Z}} &= \mathcal{\hat{E}_\text{causal}}({\hat{Q}} \oplus {\hat{Z_p}})  \\
        {\hat{I}} &= \mathcal{D}(\mathcal{R}(\tilde{Q}_{[-k:]} \oplus \tilde{Z}_{[:-k]}))
\end{split}\label{equ:decoding}
\end{equation}

\subsubsection{Training objective}

For tokenizer optimization, we adopt the training losses used in VQGAN~\cite{vqgan,llamagen}, except for the slight difference in the vector quantization loss $\mathcal{L}_{vq}$. Given that we use two separate codebooks to quantize the holistic and local patch features independently, the vector quantization loss $\mathcal{L}_{vq}$ can be formulated as $\mathcal{L}_{vq}= \mathcal{L}_{vq}(\overline{Q}) +  \mathcal{L}_{vq}(\overline{Z_p})$. Consequently, the total losses for guiding the training of our image tokenizer can be represented as $\mathcal{L} = \alpha \cdot \mathcal{L}_{vq} + \lambda \cdot \mathcal{L}_{AE}$. In this case, we set both $\alpha$ and $\lambda$ to 1.

\subsection{Autoreggresive Image Generation}

Once the tokenizer is trained, a standard autoregressive (AR) transformer model, \textit{e.g.} Llama~\cite{touvron2023llama}, can be seamlessly integrated without requiring any revision to generate these latent tokens sequentially. Starting from the class/text embedding $c$, the AR model first generates holistic tokens as prefix prompts, then guides the subsequent generation of patch token sequence in the way of next-token prediction. Once all the visual tokens are generated, they undergo the token fusion and decoding procedure to be converted into pixels. Additionally, in AR models, each visual token receives a positional embedding. Since holistic tokens form a 1D sequence, we add learnable positional embeddings for them. Meanwhile, image patch tokens continue to use 2D RoPE~\cite{rope} for their positional embeddings, which remains the same as LlamaGen~\cite{llamagen}

\subsection{Discussions}
\noindent\textbf{What do holistic tokens capture?} To understand what has been learned by the holistic tokens, we perform an experiment on the learned holistic tokens. Specifically, two images are first encoded and quantized into holistic tokens and image patch tokens, respectively. Then, the holistic tokens from one image are concatenated with the patch tokens from the other. These tokens are fed through the de-quantization process and the decoder to synthesize a new image. This entire procedure is conducted using a pre-trained holistic tokenizer, without any additional model training or fine-tuning. As shown in Fig.~\ref{fig:primary-figure}\textcolor{iccvblue}{a}, the holistic tokens can capture global information from an image, such as texture, color, materials, shapes, and other features from the pixels. For instance, in the case of the lizard image, {Hita} successfully transfers the material properties of the ship from the reference image to the input image, creating a realistic mechanical lizard. More samples are shown in Appendix.

\vspace{0.05in}\noindent\textbf{Does holistic tokenizer enhance semantic coherence?}
To verify whether the holistic features can enhance the global coherence of the generated content, we perform a zero-shot image in-painting evaluation. In this in-painting task, the model is forced to complete the lower half of an image given only the upper half. This setup prevents local information leakage during in-painting. We first tokenize the incompleted images into holistic and patch tokens, then feed them to the AR model, conditioned on class embeddings, to generate additional tokens. Hita-XXL is used without modifications to the model architecture or parameter tuning. 

As shown in Fig.~\ref{fig:primary-figure}\textcolor{iccvblue}{b}, {Hita} effectively completes the lower part of the image based on the upper part, producing images that maintain strong semantic consistency. In contrast, outputs from LlamaGen \cite{llamagen} lack this consistency. For example, in the case of the fish image, {Hita} correctly generates a complete fish, while LlamaGen~\cite{llamagen} mistakenly generates a “fish-bird”. These results confirm that our holistic tokenizer is semantic-aware and has a stronger ability to better generate semantically and holistically coherent examples. Furthermore, the ablation study of linear probing on holistic tokens is conducted in Sec.~\ref{exp:linear_prob}.

\vspace{-4pt}
\section{Experiments}

\begin{table*}[htbp]
	\centering
	\begin{tabular}{c|l|c|c|c|ccccc}
		\toprule
  Type & Model & \#Parameters & \#Tokens& \#Epochs & FID$\downarrow$ & IS$\uparrow$ & Precision$\uparrow$ & Recall$\uparrow$ \\
  \hline
  \multirow{2}*{GAN} & GigaGan~\cite{kang2023scaling} & 569M & --& -- & 3.45 & 225.5 & 0.84 & 0.61 \\
  ~ & StyleGan-XL~\cite{sauerscaling} & 166M & --& -- & 2.30 & 265.1 & 0.78 & 0.53 \\
    \hline
    \multirow{2}*{Diff.} & LDM-4~\cite{sd2.0} & 400M & -- &  -- & 3.60 & 247.7 & 0.87 & 0.48 \\
    ~ & DiT-XL/2~\cite{peebles2023scalable} & 675M & -- & -- &  2.27 & 278.2 & 0.83 & 0.57 \\
\hline
\multirow{2}*{Mask.} & MaskGIT~\cite{maskgit} & \multirow{2}{*}{227M} &  \multirow{2}{*}{256} & -- & 6.18 & 182.1 & 0.80 & 0.51 \\
~ & MaskGIT-re~\cite{maskgit} &  &  & -- & 4.02 & 355.6 & -- & -- \\
\hline
\multirow{20}*{AR} & TiTok-Base~\cite{titok}$^{\dagger}$ & 111M & \multirow{2}{*}{256} & \multirow{2}{*}{50} & 9.19 & 170.4 & 0.85 & 0.33 \\
~ & TiTok-Large~\cite{titok}$^{\dagger}$ & 343M &  &  & 5.66 & 206.4 & 0.84 & 0.45 \\
~ & GPT2-re~\cite{vqgan} & 1.4B & 256 & -- & 5.20 & 280.3 & -- & -- \\
~ & VIM-L-re~\cite{vit-vqgan} & 1.7B & 1024 & -- & 3.04 & 227.4 & -- & -- \\
~ & RQTran.-re~\cite{residual-vq} & 3.8B & 68 & -- & 3.80 & 323.7 & -- & -- \\ 

\cline{2-9}
~ & LlamaGen-B~\cite{llamagen} & 111M & \multirow{5}{*}{576} &\multirow{5}{*}{50} & 8.31 & 154.7 & 0.84 & 0.38 \\
  ~ & LlamaGen-L~\cite{llamagen}  & 343M &  & & 4.24 & 206.7 & 0.83 & 0.49 \\
 ~ & LlamaGen-XL~\cite{llamagen}  & 775M &  & & 3.24 & 245.7 & 0.83 & 0.53 \\
 ~ & LlamaGen-XXL~\cite{llamagen} & 1.4B &  & & 2.89 & 236.2 & 0.81 & 0.56 \\
 ~ & LlamaGen-3B~\cite{llamagen} & 3B &  & & 2.61 & 251.9 & 0.80 & 0.56 \\
\cline{2-9}
~ & Hita-B & 111M & \multirow{5}{*}{569} & \multirow{5}{*}{50} & 5.85 & 212.3 & 0.84 & 0.41 \\
~ & Hita-L  & 343M &  & & 3.75 & 262.1 & 0.85 & 0.48 \\
~ & Hita-XL  & 775M &  & & 2.98 & 253.4  & 0.84 & 0.54 \\
~ & Hita-XXL & 1.4B &  & & {2.70} & {274.8} & {0.84} & 0.55 \\
~ & Hita-2B & 2B &  & & \textbf{2.59} & \textbf{281.9} & \textbf{0.84} & \textbf{0.56} \\
\cline{2-9}
~ & Titok-B$^{\dagger}$~\cite{titok} & \multirow{3}{*}{111M} & 256 & \multirow{3}{*}{300} & 6.91 & 173.3 & 0.83 & 0.42 \\
~ & LlamaGen-B~\cite{llamagen}  &  & 576 & & 6.09 & 182.5  & 0.85 & 0.42 \\
~ & Hita-B &  & 569 & & {4.33} & {238.9} & {0.85} & 0.48 \\
\cline{2-9}
~ & Titok-L$^{\dagger}$~\cite{titok}  & \multirow{3}{*}{343M} & 256 & \multirow{3}{*}{300} & 4.00 & 242.1 & 0.84 & 0.50 \\
~ & LlamaGen-L~\cite{llamagen} &  & 576 & & {3.07} & {256.1} & {0.83} & 0.52 \\
~ & Hita-L &  & 569 & & {2.86} & {267.3} & {0.84} & 0.54 \\
		\bottomrule
	\end{tabular}
    \vspace{-0.6pc}
	\caption{Class-conditional image generation quality on Imagenet~\cite{imagenet} benchmark. $^{\dagger}$ indicates the approach is implemented by us `-re' indicates using rejection sampling.}
	\label{tbl:generation}
	\vspace{-1.5pc}
\end{table*}

\subsection{Setup}

\paragraph{Image Tokenizer.}

Following \cite{llamagen, titok}, we set the codebook vector dimension of the patch-level image token and holistic token to 8 and 12, respectively. As discussed in \cite{llamagen}, this design can achieve a better reconstruction and efficient codebook usage,  Besides, both codebooks have a size of $16, 384$. The number of queries for holistic feature capture is fixed at 128. Moreover, to preserve the tokenizer's simplicity, the depth of each transformer is set to 3, and the holistic token selection length $k$ is set to 4, by default.

Hita is optimized on the ImageNet~\cite{imagenet} training set and evaluated on the validation set. To ensure a fair comparison with LlamaGen~\cite{llamagen}, we train the tokenizer on $336\times336$ images, which yields a comparable number of discrete tokens to it. We also adopt the tokenizer training settings from LlamaGen. Additionally, during evaluation, the test images are resized to $256\times256$, consistent with the evaluation procedure in LlamaGen.

\vspace{-1pc}
\paragraph{Class-conditional Autoregressive Image Generation.} 
Following the AR generation procedure in LlamaGen~\cite{llamagen}, the AR models first generate images of the same size as the image tokenizer and then resize them to $256\times256$ for evaluation. Unless otherwise specified, all models are trained with the same settings as LlamaGen, with the entire training process lasting 50 epochs. Additionally, classifier-free guidance~\cite{cfg} is used in the AR generation process.

\vspace{-1pc}
\paragraph{Evaluation metrics.} 

To estimate image generation performance, we use Fréchet inception distance (FID)\cite{fid} and Inception Score (IS)\cite{is} as the main metrics to measure the generation quality of different models. In addition, Precision and Recall\cite{prec_recall} are also reported as secondary metrics.

\subsection{Main Results}

\vspace{-0.3pc}\noindent\textbf{Class-conditional Image Generation.} In line with LlamaGen~\cite{llamagen}, we evaluate our autoregressive generation models with parameters of 111M (Hita-B), 343M(Hita-L), 775M (Hita-XL) and 1.4B (Hita-XXL) on ImageNet~\cite{imagenet} $256\times256$ class-conditional generation task, and compare them with the mainstream generation model families, including generative adversarial networks (GAN)~\cite{kang2023scaling, sauerscaling}, diffusion models (Diff.)~\cite{dhariwal2021diffusion, ho2022cascaded, sd2.0, peebles2023scalable}, BERT-style masked-prediction models (Mask.)~\cite{maskgit}, and AR generation models~\cite{titok, vqgan, vit-vqgan, residual-vq, vqvae-2, llamagen}. 

\begin{figure*}[thbp]
  \centering
  \includegraphics[width=1\textwidth]{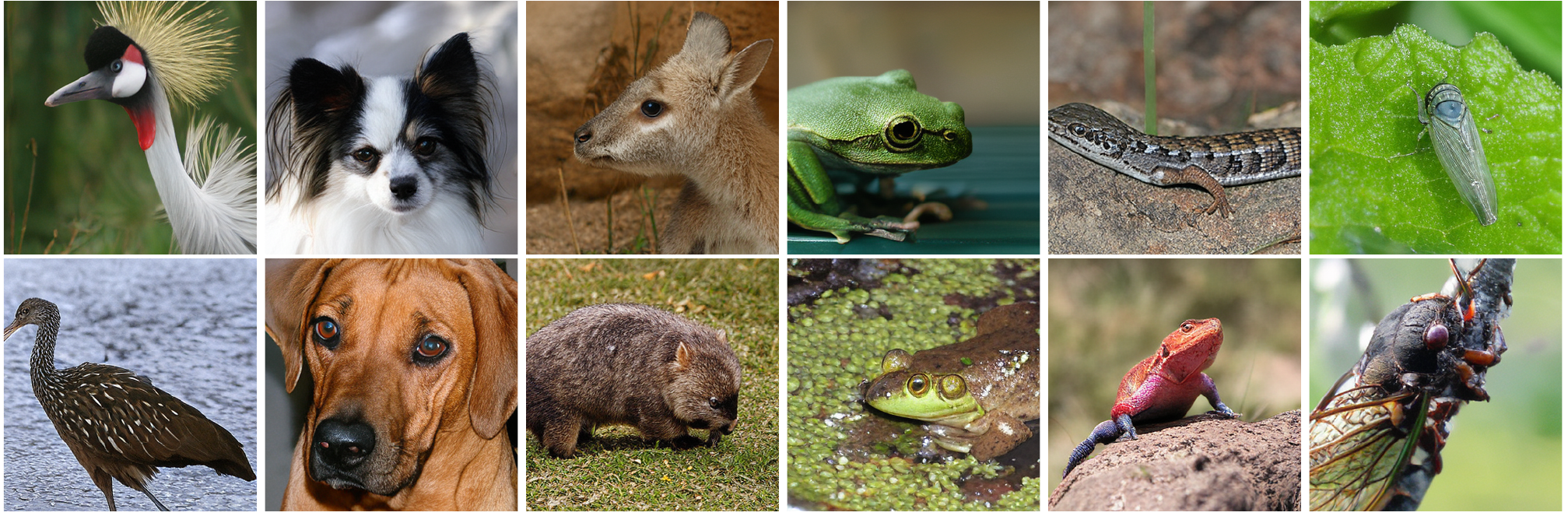}
  \vspace{-1.7pc}
\caption{Class-conditional image generation with our proposed Hita.}
\label{fig:conditional-img-gen}
\vspace{-1.4pc}
\end{figure*}

As shown in Table~\ref{tbl:generation}, our models exhibit competitive performance across all metrics compared to mainstream image generation models. Notably, our model outperforms popular diffusion models LDM like LDM-4\cite{sd2.0} by approximately~\textbf{0.7} FID and \textbf{19.6} IS with~\textbf{16.6\%} fewer parameters. Analogously, {Hita} beats BERT-style models~\cite{maskgit} in terms of FID without the requirement of complicated sampling tuning. With comparable or even fewer parameters, our method surpasses most AR generative models~\cite{titok, vqgan, vit-vqgan, residual-vq, vqvae-2} in both FID and IS metrics. Under the same training setting, Hita surpasses LlamaGen~\cite{llamagen} by significant FID gains and notable IS improvements. Beside, as shown in the table, extending the training duration leads to notable performance improvements of our approach over the baseline.~\textit{e.g.}, Hita-B outperforms LlamaGen-B~\cite{llamagen} with gains of \textbf{1.76} in FID and \textbf{56.4} in IS. Besides, our Hita-L model achieves an FID of \textbf{2.86} at 300 epochs. Beyond longer training duration, we also train an AR generation model with 2B parameters for 50 epochs. As depicted in Table.~\ref{tbl:generation}, our Hita-2B exhibits competitive generation performance and even surpasses LlamaGen-3B~\cite{llamagen} with much fewer parameters. Additionally, class-conditional image generation results are also presented in Fig.~\ref{fig:conditional-img-gen}.

\vspace{-0.5pc}
\subsection{Ablation Study}


First, we study the impact of learnable queries and semantic injection. Second, we analyze the selection length $k$. Next, we  discuss different foundation models for semantic injection. Later, we study the semantic level of holistic tokens. We provide further ablation in the Appendix.

\vspace{0.03in}\noindent\textbf{Learnable queries and Semantic Injection Analysis.} 
In {Hita}, it incorporates a set of learnable queries and leverages semantic injection from a pre-trained model (\textit{e.g.}, DINOv2~\cite{dinov2reg}) to capture holistic properties. To validate their effectiveness, we conduct analysis on the learnable queries ${Q}$ and semantic injection $\mathcal{H}$. Specifically, We first train a tokenizer with only the attention modules as a baseline. Then, we gradually add learnable queries and a pre-trained foundation model (e.g., DINOv2) to this baseline.  Next, we train a Hita-B AR generation model for 50 epochs to estimate the generation quality. Moreover, we conduct linear probing on the whole latent tokens after the last attention module in the baseline. As for the comparison experiments, we only perform linear probing on the holistic tokens from the last attention module. As depicted in Table.~\ref{tbl:ablation_queries}, simply introducing learnable queries can significantly improve image reconstruction and generation quality. The additional features from the foundation model can further enhance the reconstruction and generation quality. Besides, as indicated by the linear probing, the learnable queries can capture better semantic information than the baseline. With semantics injected, this semantic-level representation can be further enhanced. This illustrates that combining the learnable queries and semantic injection is essential for learning better holistic information, as lacking either degrades image reconstruction and generation quality. 

\begin{table}[ht]
	\centering
    \vspace{-0.3cm}
    \setlength{\tabcolsep}{.8pt}
	\begin{tabular}{c|c|cc|cc|cc|c}
		\toprule
    \multirow{2}{*}{$\textit{setup}$} & \multicolumn{3}{c|}{$\textit{image recon.}$} &\multicolumn{2}{c|}{usage(\%) $\uparrow$} & \multicolumn{2}{c|}{$\textit{AR gen.}$} & \multirow{2}{*}{L.P} \\
    \cline{2-8}
    & \#toks & rFID$\downarrow$ & rIS$\uparrow$ & $\mathcal{Q}_{H}$ & $\mathcal{Q}_{P}$ & gFID$\downarrow$ & gIS$\uparrow$ & \\
    \hline
    Baseline & 441 &  1.31 & 193.3 & -- & 100.0 & 9.37 & 162.6 & 14.2 \\
    \hline
    +Queries & \multirow{2}{*}{569} & 1.15 & 191.9 & 30.2 & 100.0 & 6.32 & 187.9 & 28.2 \\
    + DINOv2 &  & \textbf{1.03} & \textbf{198.5}  & 100.0 & 100.0 & \textbf{5.85} & \textbf{212.3} & 36.6 \\
    \bottomrule
	\end{tabular}
    \vspace{-0.5pc}
    \caption{Ablation of learnable and semantic injection from pre-trained models. The learnable queries and DINOv2~\cite{dinov2reg} feature are gradually added to the baseline model. L.P. indicates linear probing on the ImageNet~\cite{imagenet} validation set.} 
	\label{tbl:ablation_queries}
	\vspace{-1.3pc}
\end{table}

\vspace{0.05in}\noindent\textbf{Is $k > 0$ crucial?} 
To emphasize holistic tokens and avoid its codebook collapse, {Hita} first builds a 1D ordered sequence with holistic tokens first and followed by the flattened patch-level tokens. Next, it is fed into a causal transformer for token fusion. Subsequently, the last $k$ holistic tokens are combined with patch-level tokens to reconstruct the image. To verify the doubt, we first initialize and train the holistic tokenizer, configuring $k$ with different values. Next, we train an AR generation model -- Hita-L with fixed 128 holistic tokens in total. Consequently, both image reconstruction and generation quality are estimated. As listed in Table~\ref{tbl:ablation_k}, when $k = 0$, the token fusion module degrades to a vanilla causal transformer. In this case, the patch-level tokens can directly influence the reconstruction of their corresponding patches, creating a shortcut that bypasses the holistic tokens. As a result, the token fusion process ignores the holistic tokens, leading to its codebook collapse, as evidenced by the 2nd row in Table.~\ref{tbl:ablation_k}.  Once $k > 0$, the patch-level tokens can only affect the reconstruction of other patches and must also rely on the holistic tokens. In this scenario, the problem of holistic codebook collapse is avoided. This also improves the codebook utilization and the reconstruction quality. Here, we also found that $k$ equals 4 achieves the optimal reconstruction and generation quality. Thus, we set $k$ to 4 by default in our experiments.

\vspace{-1pc}
\begin{table}[ht]
	\centering
    \setlength{\tabcolsep}{4.2pt}
	\begin{tabular}{c|cc|cc|cccc}
		\toprule
            \multirow{2}{*}{${k}$} & \multicolumn{2}{c|}{$\textit{Image recon.}$} &\multicolumn{2}{c|}{\textit{code usage}$\uparrow$} & \multicolumn{2}{c}{$\textit{AR gen.}$} \\
            \cline{2-7}
             & rFID$\downarrow$ & rIS$\uparrow$ & $\mathcal{Q}_{H}$ & $\mathcal{Q}_{P}$ & gFID$\downarrow$ & gIS$\uparrow$   \\
            \hline
            0 & 1.34 & 187.2 & 0.763\% & 100.0\% & 6.80 & 145.4 \\
            1 & 1.13 & 192.4 & 100.0\% & 100.0\%  & 3.78 & 257.1  \\
            2 & 1.08 & 195.1& 100.0\% & 100.0\%  & 4.14 & 256.8  \\
            4 & \textbf{1.03} & \textbf{198.5} & 100.0\% & 100.0\% & \textbf{3.75} & \textbf{268.8}\\
		\bottomrule
	\end{tabular}
    \vspace{-.4pc}
	\caption{Ablation study of the selection length $k$ of holistic tokens.}
	\label{tbl:ablation_k}
	\vspace{-1.2pc}
\end{table}

\vspace{0.03in}\noindent\textbf{Pre-trained models for semantic injection.} When generating holistic features, a pre-trained foundation model (e.g., DINOv2~\cite{dinov2reg}) is adopted to inject semantic information. However, the impact of using different pre-trained models for semantic injection on image reconstruction and generation remains unexplored. Here we study injecting semantic-aware features from different pre-trained models, including the vision language model (e.g., CLIP~\cite{clip}) or vanilla vision backbone (e.g., ResNet~\cite{resnet}). We initialize and train a holistic tokenizer with different pre-trained foundation models using the default settings. Then we train Hita-L for default 50 epochs. As shown in Table~\ref{tbl:pretrained_model}, injecting DINOv2's semantic features yields better results. Therefore, we utilize DINOv2 to perform semantic injection as default. 

\vspace{-1pc}
\begin{table}[ht]
	\centering
        \setlength{\tabcolsep}{0.8pt}
	\begin{tabular}{c|cc|cc}
		\toprule
            \multirow{2}{*}{$\textit{model}$} & \multicolumn{2}{c|}{$\textit{ image recon.}$} & \multicolumn{2}{c}{$\textit{AR gen.}$} \\
            \cline{2-5}
             & rFID$\downarrow$ & rIS$\uparrow$  & gFID$\downarrow$ & gIS$\uparrow$   \\
            \hline
            None & 1.15 & 191.9 & 6.32 & 187.9   \\
            CLIP~\cite{clip} & 1.24 & 191.9  & 6.42 & 186.3  \\
            ResNet~\cite{resnet} & 1.07 & 195.3  & 6.81 & 191.9 \\
            DINOv2~\cite{dinov2reg} & \textbf{1.03} & \textbf{198.5} & \textbf{5.85} & \textbf{212.3} \\
		\bottomrule
	\end{tabular}
    \vspace{-0.6pc}
	\caption{Comparison of semantic injection performance with different pre-trained models.}
	\label{tbl:pretrained_model}
	\vspace{-1.6pc}
\end{table}

\vspace{0.1in}
\noindent\textbf{What is the semantic-level of latent tokens?} 
\label{exp:linear_prob}
To analyze the semantic-level of the learned holistic tokens, we conduct linear probing according to the setting of MAE~\cite{mae} on ImageNet~\cite{imagenet} validation set. Specifically, the holistic tokens from the final causal transformer are used. Here we focus on estimation rather than engaging in a comprehensive comparison with all alternative approaches.

Compared to the vanilla tokenizer VQGAN~\cite{llamagen}, we notice that Hita achieves a better accuracy, suggesting it captures semantic information. Nevertheless, it lags behind DINOv2 ~\cite{dinov2reg}. Because the semantic features learned by Hita are different from DINOv2's. As shown in Fig.~\ref{fig:primary-figure}\textcolor{iccvblue}{a}, Hita's primary goal is image reconstruction, focusing more on learning relatively low-level global features like texture, material, color, and other features from pixels. Nevertheless, DINOv2, a pre-trained model integrated in Hita, injects necessary semantic-aware features to aid reconstruction. Conversely, DINOv2 tends towards understanding image content, with self-supervised ViT features containing explicit information about the semantic segmentation of an image, capturing more general and high-level features.

\vspace{-0.6pc}
\begin{table}[ht]
	\centering
	\begin{tabular}{c|c|c|c}
		\toprule
		   & VQGAN~\cite{llamagen}& Hita & DINOv2~\cite{dinov2reg}  \\
            \hline
            Acc.@Top-1 & 6.9 & 36.6 & 86.7 \\
		\bottomrule
	\end{tabular}
    \vspace{-.4pc}
	\caption{Linear probing accuracy comparison on ImageNet \cite{imagenet}.}
	\label{tbl:extending_duration}
	\vspace{-1.65pc}
\end{table}

\section{Conclusion}
Autoregressive (AR) models struggle with image generation because their unidirectional attention mechanisms restrict their ability to capture global context, resulting in lower generation quality. Hita addresses this by introducing a holistic-to-local tokenization scheme that uses learnable holistic tokens and local patch tokens to incoporate global information. Beyond image tokenization, {Hita} exhibits new features such as zero-shot style transfer and zeros-shot image in-painting. Experiments conducted on class-conditional image generation tasks illustrate that it accelerates convergence speed during training and significantly improves the synthesis quality. Besides, in-depth analysis also consolidates its effectiveness.

\section*{Acknowledgments}
This work has been supported by the National Key R\&D Program of China (Grant No. 2022YFB3608300), Hong Kong Research Grant Council - Early Career Scheme (Grant No. 27209621), General Research Fund Scheme (Grant No. 17202422, 17212923, 17215025) Theme-based Research (Grant No. T45-701/22-R) and Shenzhen Science and Technology Innovation Commission (SGDX20220530111405040). Part of the described research work is conducted in the JC STEM Lab of Robotics for Soft Materials funded by The Hong Kong Jockey Club Charities Trust.  We are deeply grateful to Lufan Ma for the contribution in editing this paper.

{
    \small
    \bibliographystyle{ieeenat_fullname}
    \bibliography{main}
}
\clearpage

\appendix

First, we compare Hita with other vanilla tokenizers and further discuss the token fusion module. Then, we elaborate on more ablations in detail. Later, we present more visualization samples egarding zero-shot style transfer, zero-shot in-painting, and class-conditional image generation.

\vspace{-0.6pc}
\section{Comparison with other image tokenizers.}
In this subsection, we first compare Hita with the other vanilla tokenizers. Next, we depict the differences between VAR~\cite{var} in detail. Next, we present further discussion on token fusion modules.

\vspace{-0.6pc}
\subsection{Comparison with vanilla image tokenizers}

In Table.~\ref{tbl:tokenizer_comparison}, we compare with the prevalent image tokenizer, including VQGAN~\cite{vqgan}, MaskGiT~\cite{maskgit}, ViT-VQGAN~\cite{vit-vqgan} and TiTok~\cite{titok}. The image reconstruction quality is measured by rFID~\cite{fid} and rIS~\cite{is} metrics, which are evaluated on 256$\times$256 ImageNet~\cite{imagenet} 50k validation benchmark. 

Following VQGAN~\cite{vqgan}, we adopted ${\ell}_2$-normalization into codebook vectors, low codebook vector dimension, and a codebook size of 16,384. Compared with other counterparts like VQGAN~\cite{vqgan} and ViT-VQGAN~\cite{vit-vqgan}, the proposed tokenizer represents an image with fewer tokens (569 \textit{vs.} 1024), while achieving a better reconstruction quality with $\textbf{100\%}$ utilization for both holistic and patch-level codebooks. Additionally, we observed that our approach achieves a better rIS \textbf{198.5} compared with the VQGAN proposed in LlamaGen~\cite{llamagen}. rIS quantifies the KL-divergence between the original label distribution and the logit distribution of reconstructed images after softmax normalization~\cite{is}. In other words, rIS measures the semantic consistency between the reconstructed images and the original ones. A  higher rIS confirms that our holistic tokenizer is more effective at preserving the semantic consistency of the reconstructed images.
\begin{table}[htbp]
	\centering\small
    \setlength{\tabcolsep}{1.1pt}
	\begin{tabular}{l|c|ccc|cc|cc}
		\toprule
    \multirow{2}{*}{Approach} & \multicolumn{1}{c|}{$\textit{f}$} &\multicolumn{3}{c|}{\textit{setup}} & \multicolumn{2}{c|}{$\textit{img recon.}$} & \multicolumn{2}{c}{\textit{usage(\%)$\uparrow$}} \\
    \cline{3-9}
     &  & size & dim & \#toks & rFID$\downarrow$ & rIS $\uparrow$ & $\mathcal{Q}_{H}$ & $\mathcal{Q}_{P}$ \\
    \hline
    TiTok~\cite{titok} & -- & 8,192 & 64 & 256 & 1.05 & 191.5 
    & -- & 100.0 \\
    \hline
    $\text{VQGAN}^{\text{oim.}}$~\cite{vqgan} & \multirow{4}{*}{8} & 256 & 4 & \multirow{4}{*}{1024} & 1.44 & -- & -- & -- \\
    VQGAN~\cite{vqgan} & ~ & 8192 & 256 & ~ & 1.49 & -- & -- & -- \\
    ViT-VQGAN~\cite{vit-vqgan} & ~ & 8192 & 32 & ~ & 1.28 & 192.3 & -- & 95.0  \\
    $\text{VQGAN}^{\text{oim.}}$~\cite{vqgan} & ~ & 16384 & 4 & ~ & 1.19 & -- & -- & --\\
    \hline
    VQGAN~\cite{vqgan} & \multirow{2}{*}{16} & \multirow{2}{*}{1024} & \multirow{2}{*}{256} & \multirow{2}{*}{256} & 7.94 & -- & -- & -- \\
    MaskGiT~\cite{maskgit} & ~ & ~ & ~ & ~ & 2.28 & -- & -- & -- \\
    \hline
    Var~\cite{var} & 16 & 4096 & 32 & 680 & \textbf{0.92} & 196.0 & -- & 100.0 \\
    \hline
    RQ-VAE~\cite{residual-vq} & 32 & 16384 & 256 & 1024 & 1.83 & -- & -- & -- \\
    \hline
    VQGAN~\cite{vqgan} & \multirow{4}{*}{16} & \multirow{4}{*}{16384} & 256 & 256 & 4.98 & -- & --  & --\\
    VQGAN~\cite{llamagen} & ~ & ~ & 8 & 441 & 1.21 & 189.1 & -- & 99.2 \\
    VQGAN~\cite{llamagen} & ~ & ~ & 8 & 576 & {0.95} & 197.3 & -- & 99.7 \\
    Hita & ~ & ~ & 8/12 & 569 & 1.03 & \textbf{198.5} & 100.0 & 100.0 \\
 		\bottomrule
	\end{tabular}
    \vspace{-0.5pc}
    \caption{Comparison with other image tokenizers. $^\text{oim.}$ indicates training on OpenImages~\cite{openimage}. $\mathcal{Q}_{H}$/$\mathcal{Q}_{P}$ denote the codebook usage in holistic and patch-level quantizers, respectively.}
	\label{tbl:tokenizer_comparison}
	\vspace{-0.5pc}
\end{table}

\vspace{-0.2pc}
\subsection{Comparison with VAR}
As discussed in Sec.\textcolor{iccvblue}{2}, VAR's multi-scale tokens can also be considered as a combination of semantic tokens and patch tokens. However, as depicted in Table.~\ref{tbl:abl_var} we find removing the initial coarse-scale tokens seldom effects its reconstruction. Meanwhile, linear probing conducted on the cumulative coarse-scale tokens reveals poor semantic information.

\begin{table}[h]
\centering
\setlength{\tabcolsep}{4.2pt}
\scalebox{0.75}{
	\begin{tabular}{c|c|c|c|c|c|c|c|c|c}
		\toprule
$n$ & 0 & 1 & 2 & 3 & 4 & 5 & 6 & 7 & 8 \\
\hline
rFID & 1.31 & 1.31 & 1.32 & 1.31 & 1.41 & 1.78 & 3.22 & 10.42 & 92.3 \\
\hline
rIS & 198.6 & 199.3 & 198.8 & 198.9 & 196.4 & 190.2 & 171.8 & 119.8 & 16.4 \\
\hline
Acc & 2.2 & 4.9 & 7.2 & 8.3 & 9.2 & 9.5 & 9.8 & 10.1 & 10.3 \\
		\bottomrule
	\end{tabular}}
    \vspace{-0.4pc}
    \caption{Analysis of VAR's multi-scale tokens. Acc indicates top-1 accuracy for linear probing estimation on the ImageNet~\cite{imagenet}.}
	\label{tbl:abl_var}
    \vspace{-1.3pc}
\end{table}

\section{Explanation on Token Fusion Module}
\vspace{-0.2pc}
\begin{figure}[htbp]
  \centering
  \includegraphics[width=.48\textwidth]{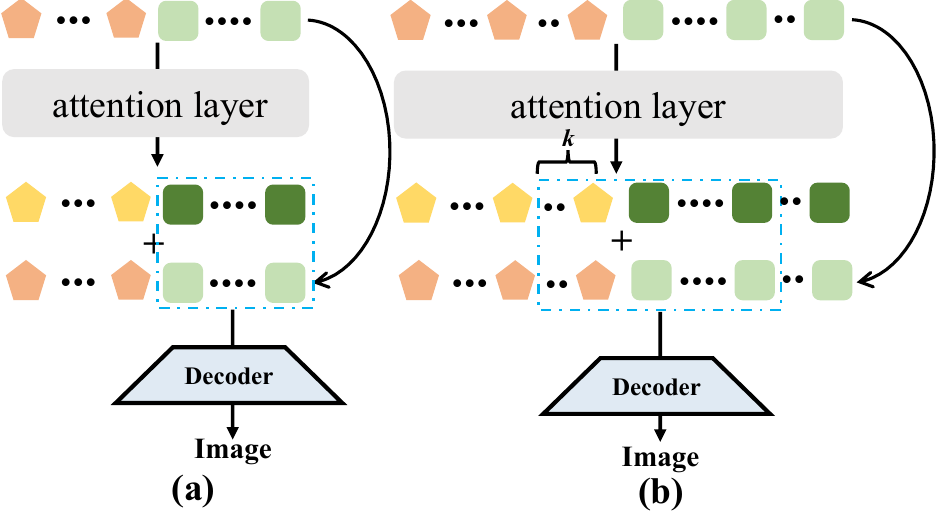}
  \vspace{-1.6pc}
\caption{Token fusion module comparison. (a).Token fusion module composed of vanilla attention layers. (b). Our designed token fusion layers.}
\label{fig:skip-connection}
\vspace{-1.0pc}
\end{figure}

As shown in Fig.~\ref{fig:skip-connection}, If a vanilla transformer is used to construct the token fusion module, due to the existence of the skip connection and the patch-level tokens contain enough information for image reconstruction, the patch-level tokens can directly flow through the skip connections into the decoder to reconstruct the image. Thus, the token fusion module can learn a trivial solution, which overlooks the holistic tokens and leads to holistic codebook collapse (Fig.~\ref{fig:skip-connection}\textcolor{iccvblue}{a}). Once the last $k$ holistic tokens participate in the image reconstruction, the information of the patch tokens flowing through the skip connection is incomplete. It needs to interact with the holistic tokens to obtain complete information for image reconstruction(Fig.~\ref{fig:skip-connection}\textcolor{iccvblue}{b}). This simple operation emphasizes the holistic tokens and avoids codebook collapse. To better align the token sequence with the nature of the AR generation model, here we adopt causal attention to build the token fusion module.

\vspace{-0.5pc}
\section{More Ablation Studies.}
We elaborate on further ablation studies on the design of our approach. Next, we quantitatively analyze Hita’s zero-shot inpainting performance.

\begin{figure*}[htbp]
  \centering
  \includegraphics[width=1\textwidth]{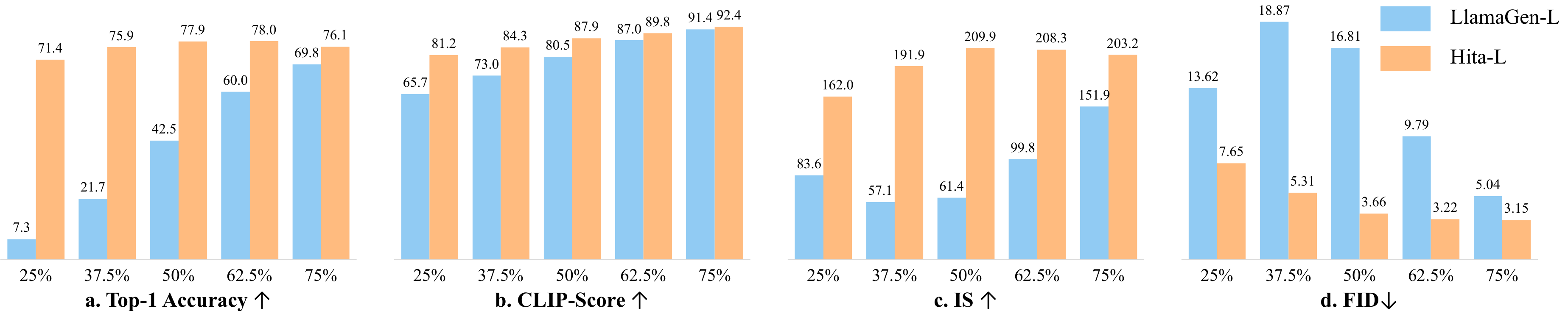}
\vspace{-1.25pc}
\caption{Quantitative zero-shot in-painting analysis with Hita-L conducted on ImageNet~\cite{imagenet} evaluation benchmark.}
\label{fig:quantitative-in-painting}
\vspace{-1pc}
\end{figure*}

\subsection{AR generation with resolution of \textbf{$256\times256$}}

Given that Hita ensures a fair comparison with other approaches by controlling the number of tokens, here we directly train the image tokenizer and the AR generation model on $256\times256$ images, enabling them to reconstruct and generate $256\times256$ images, respectively, which is also in line with common practice. We initialize and train the image tokenizer for 40 epochs, and train both Hita-B and Hita-L for 50 epochs as default. As depicted in Table.~\ref{tbl:low_resolution}, Hita can not only achieve a better reconstruction performance, but also improve the generation quantity compare to LlamaGen~\cite{llamagen}.

\begin{table}[ht]
	\centering
    \setlength{\tabcolsep}{2.0pt}
	\begin{tabular}{l|cc|cc|cccc}
		\toprule
            \multirow{2}{*}{Approach} & \multicolumn{2}{c|}{$\textit{Image recon.}$} &\multicolumn{2}{c|}{\textit{code usage}$\uparrow$} & \multicolumn{2}{c}{$\textit{AR gen.}$} \\
            \cline{2-7}
            & rFID$\downarrow$ & rIS$\uparrow$ & $\mathcal{Q}_{H}$ & $\mathcal{Q}_{P}$ & gFID$\downarrow$ & gIS$\uparrow$   \\
            \hline
            LlamaGen-B  & \multirow{2}{*}{2.22} &\multirow{2}{*}{169.8} & \multirow{2}{*}{--} & \multirow{2}{*}{95.2\%}  & 7.22 & 178.3  \\
            LlamaGen-L  &  &  &  &  & 4.21 & 200.0  \\
            \hline
            Hita-B & \multirow{2}{*}{1.40} & \multirow{2}{*}{186.6} & \multirow{2}{*}{100\%} & \multirow{2}{*}{100\%} & \textbf{6.58} & \textbf{210.2} \\
            Hita-L &  &  &  &  & \textbf{4.04} & \textbf{242.2} \\
		\bottomrule
	\end{tabular}
	\caption{Hita performs {image reconstruction} and {AR generation} with image resolution of $256\times256$.}
	\label{tbl:low_resolution}
	\vspace{-1pc}
\end{table}

\subsection{Other Token Fusion Variants}

As depicted in Sec~\textcolor{iccvblue}{3.2.3}, for simplicity, we choose the last $k$ holistic tokens combined with the patch-level tokens to reconstruct the image. Here, we refer to MAE~\cite{mae} and Titok~\cite{titok} and design 2 different variants to generate features for the first $k$ patch tokens: 1). \textbf{Partial}: A mask token combined with the positional embedding of the first $k$ patch tokens are used to generate their feature. In this scenario, the holistic tokens can be treated as condition; 2). \textbf{Full}: The patch tokens are completely removed, which is consistent with  TiTok~\cite{titok}. Then, a mask token, along with the positional embeddings, generates the features for all patch tokens. To achieve this, we initialize and train 2 different holistic tokenizers for 40 epochs, and then train 2 AR generation models -- Hita-B and Hita-L based on those tokenizers with default training settings for 50 epochs. As listed in Table~\ref{tbl:mae_style}, the token fusion strategy proposed in Hita shows a better performance in both reconstruction and generation compared to the other variants. Thus, we choose to use the last ${k}$ holistic tokens combined with the patch-level tokens to reconstruct the image, by default.

\begin{table}[ht]
	\centering\small
    \setlength{\tabcolsep}{2.2pt}
	\begin{tabular}{l|cc|cc|c|cccc}
		\toprule
            \multirow{2}{*}{Variants} & \multicolumn{2}{c|}{$\textit{image recon.}$} &\multicolumn{2}{c|}{\textit{code usage}$\uparrow$} & \multicolumn{3}{c}{$\textit{AR gen.}$} \\
            \cline{2-8}
            &  rFID$\downarrow$ & rIS$\uparrow$ & $\mathcal{Q}_{H}$ & $\mathcal{Q}_{P}$ & Model & gFID$\downarrow$ & gIS$\uparrow$   \\
            \hline
            \multirow{2}{*}{\textbf{Partial}} & \multirow{2}{*}{1.05} & \multirow{2}{*}{198.2} & \multirow{2}{*}{100.0\%} & \multirow{2}{*}{100.0\%} & B & 6.59 & 209.8 \\
             &  &  &  &  & L & 3.96 & 243.1 \\
            \hline
            \multirow{2}{*}{\textbf{Full}} & \multirow{2}{*}{2.07} & \multirow{2}{*}{170.3} & \multirow{2}{*}{100.0\%} & --  & B &  11.64 & 172.1 \\
             &  &  &  & -- & L & 6.75 & 219.7 \\
             \hline
            \multirow{2}{*}{Hita} & \multirow{2}{*}{1.03} & \multirow{2}{*}{198.5} & \multirow{2}{*}{100.0\%}  & \multirow{2}{*}{100.0\%} & B & \textbf{5.85} & \textbf{212.3} \\
             &  &  &  &  & L & \textbf{3.75} & \textbf{262.1} \\
		\bottomrule
	\end{tabular}
	\caption{{Image reconstruction} and {AR generation.} with different token fusion strategies.}
	\label{tbl:mae_style}
	\vspace{-1pc}
\end{table}

\subsection{Attention Modules Study.} As outlined in Sec.~\textcolor{iccvblue}{2}, the attention modules consists of one standard transformer $\mathcal{E}_\text{trans}(\cdot)$, two causal transformers $\mathcal{E}_\text{causal}(\cdot)$ and $\mathcal{\hat{E}_\text{causal}}(\cdot)$.  $\mathcal{E}_\text{trans}(\cdot)$ for holistic feature capture, $\mathcal{E}_\text{causal}(\cdot)$ is for causal latent space alignment, and $\mathcal{\hat{E}_\text{causal}}(\cdot)$ for holistic codebook learning. Here we study their effectiveness as follows: 1) We remove $\mathcal{E}_\text{trans}(\cdot)$ and introduce new attention mask into $\mathcal{E}_\text{causal}(\cdot)$ to learn its contribution to holistic feature capture; 2). We substitute $\mathcal{E}_\text{causal}(\cdot)$ to study its effect on causal latent space alignment. Here, we initialize and train the tokenizers with different attention modules. Then, we train an AR generation model (Hita-B) on those holistic tokenizers to estimate their generation quality. 

As depicted in Table.~\ref{tbl:component-analysis}, it can be observed when only $\mathcal{E}_\text{trans}$ is directly discarded from the tokenizer, the quality of image reconstruction and generation slightly drops. This is because the subsequent causal transformer $\mathcal{E}_\text{causal}$ can simultaneously achieve the requirement of holistic feature capture, semantic-aware feature injection, and feature space alignment. Similarly, only removing $\mathcal{E}_\text{causal}$ leads to a slight degradation in image reconstruction and generation, indicating that incorporating causal attention $\mathcal{E}_\text{trans}$ within the tokenizer helps in learning a latent space that better aligns with the causal nature of AR models. With all the modules integrated, we achieve the best performance in terms of reconstruction and generation. 

\begin{table}[ht]
	\centering
        \setlength{\tabcolsep}{0.8pt}
	\begin{tabular}{ccc|cc|cc|cc}
		\toprule
            \multicolumn{3}{c|}{$\textit{component}$} & \multicolumn{2}{c|}{$\textit{image recon.}$} & \multicolumn{2}{c|}{code usage${\uparrow}$} & \multicolumn{2}{c}{$\textit{AR gen.}$} \\
            \hline
$\mathcal{E}_\text{trans}$ & $\mathcal{E}_\text{causal}$ & $\mathcal{\hat{E}}_\text{causal}$ &  rFID$\downarrow$ & rIS$\uparrow$ & $\mathcal{Q}_{H}$ & $\mathcal{Q}_{P}$ & gFID$\downarrow$  & gIS$\uparrow$   \\
            \hline
-- & -- & -- & 1.21 & 189.1 & -- & 99.7\% & 7.95 & 166.9 \\
-- & \checkmark & \checkmark  & 1.15 & 196.0 & 100.0\% & 100.0\%  & 6.01 & 196.3 \\ 
\checkmark & -- & \checkmark  & 1.20 & 195.3 & 100.0\% & 100.0\%  & 6.12 & 188.9 \\
\checkmark & \checkmark & \checkmark & \textbf{1.03} & \textbf{198.5} & 100.0\% & 100.0\% &\textbf{5.85} & \textbf{212.3} \\ 
		\bottomrule
	\end{tabular}
    \vspace{-0.5pc}
	\caption{Attention modules analysis in Hita. `-' indicates the module was removed from the tokenizer's architecture.}
	\label{tbl:component-analysis}
	\vspace{-0.75pc}
\end{table}

\subsection{Quantitative Study of In-painting Quality.}
Beyond the demonstration that the holistic tokenizer can help AR generation model maintain a better semantic consistency (see Fig.~\ref{fig:in-painting}), we conduct a quantitative analysis of this. Similar to the experimental setup described in the manuscript, we only retain a certain fraction of upper part image, \textit{e.g.} $25\%$, $50\%$, \textit{etc.} then utilize a tokenizer to discretize it into visual tokens. These tokens are fed as a prefix sequence prompt to a pre-trained AR generation model, which is required to complete the lower part of the image. Here, we take the AR generation models -- Hita-L and LlamaGen-L~\cite{llamagen} for a fair comparison.

\vspace{0.05in}\noindent\textbf{Evaluation metrics.} 
To quantitatively estimate the semantic consistency of the generated images, we adopt top-1 accuracy and CLIP-score as our metrics, along with the generation metrics FID and IS. The top-1 accuracy is derived from image classification tasks, where we utilize a pre-trained ResNet-101~\cite{resnet} on ImageNet~\cite{imagenet} to classify the completed images. CLIP-score measures the similarity between the original and the completed images. Specifically, we feed both original and in-painted images into the CLIP~\cite{clip} model to extract image features and compute their cosine distance averaged across all samples. Higher top-1 accuracy and CLIP-scores indicate a better semantic consistency is maintained in an AR generation model.

\vspace{0.05in}\noindent\textbf{Observation and discussion.} As shown in Fig.~\ref{fig:quantitative-in-painting} and Fig.~\ref{fig:in-painting}, the AR generation model trained with vanilla VQGAN~\cite{vqgan} encounters difficulties in producing an image that maintains semantic coherence. In contrast, our approach effectively produces visually consistent content to complete the given image part, maintaining a better overall coherence even with a significantly truncated prefix. Specifically, as observed in Fig.~\ref{fig:quantitative-in-painting}.\textcolor{iccvblue}{a} and Fig.~\ref{fig:quantitative-in-painting}.\textcolor{iccvblue}{b}, our method achieves higher classification accuracy and CLIP-score under various settings, with relatively stable fluctuations in all metrics across different configurations. In contrast, models trained with vanilla VQGAN exhibit more obvious performance variations, especially when only a small portion of the image is provided. 

Beyond measuring semantic coherence, we also estimate the completed image quality using the generation evaluation metrics, FID and IS. As shown in Fig.~\ref{fig:quantitative-in-painting}.\textcolor{iccvblue}{c} and Fig.~\ref{fig:quantitative-in-painting}.\textcolor{iccvblue}{d}, compared to the model trained with vanilla VQGAN\cite{vqgan}, our approach achieves a better FID and IS, which also illustrates the holistic tokenizer can help the AR generation model generate the lower half of images
robustly, maintaining strong semantic consistency.

\section{Limitations}
Currently, Hita is trained on a limited dataset using basic optimization techniques. Better performance could be achieved with more data and advanced learning objectives. Additionally, Hita can be seamlessly extended to perform text-conditional image generation, which is an ongoing direction of our research.

\section{More Visual Examples}

\begin{figure*}[htbp]
  \centering
  \includegraphics[width=1.\textwidth]{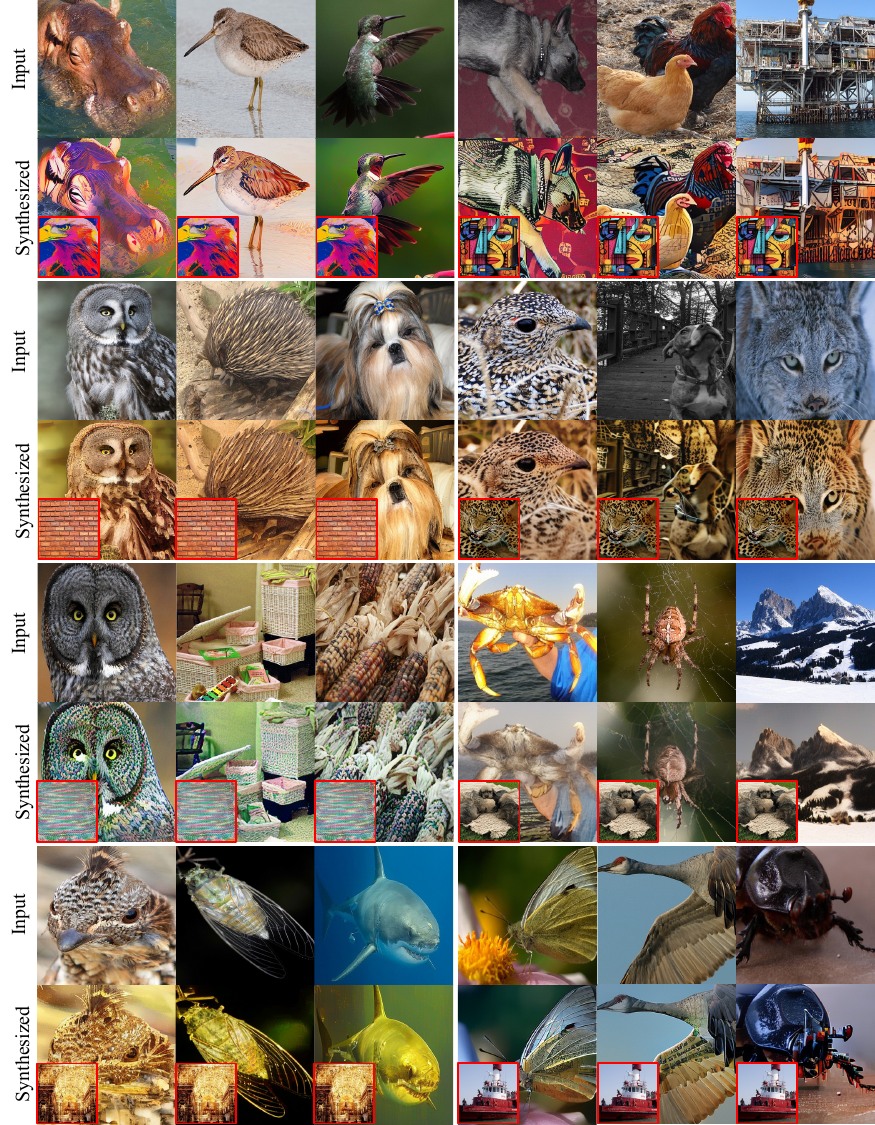}
  \vspace{-1.5pc}
\caption{Some zero-shot style-transfer samples by Hita's holistic tokenizer. Best viewed with zoom-in.
}
\label{fig:zero-shot-style-transfer}
\end{figure*}

\begin{figure*}[htbp]
  \centering
  \includegraphics[width=.95\textwidth]{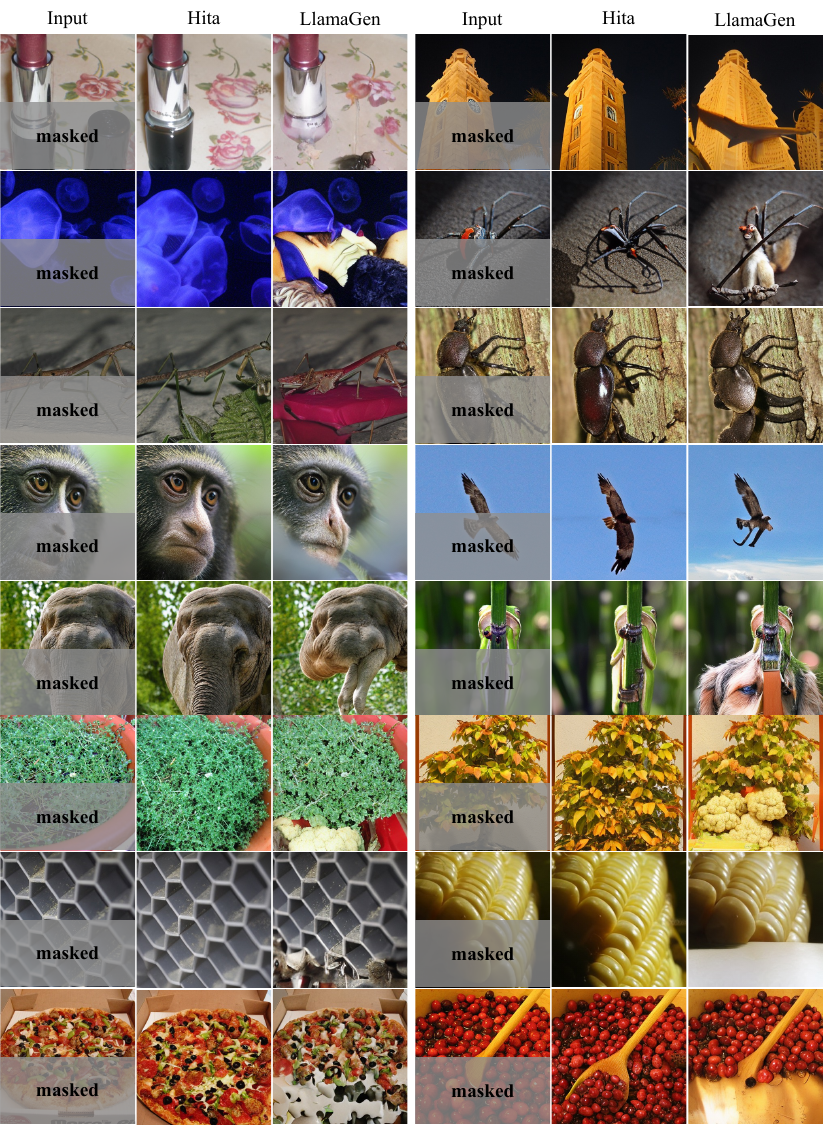}
  \vspace{-0.5pc}
\caption{Zero-shot in-painting examples by Hita's AR generation model. Compared with the baseline. Best viewed with zoom-in.}
\label{fig:in-painting}
\vspace{-2pc}
\end{figure*}

\begin{figure*}[htbp]
  \centering
  \includegraphics[width=.95\textwidth]{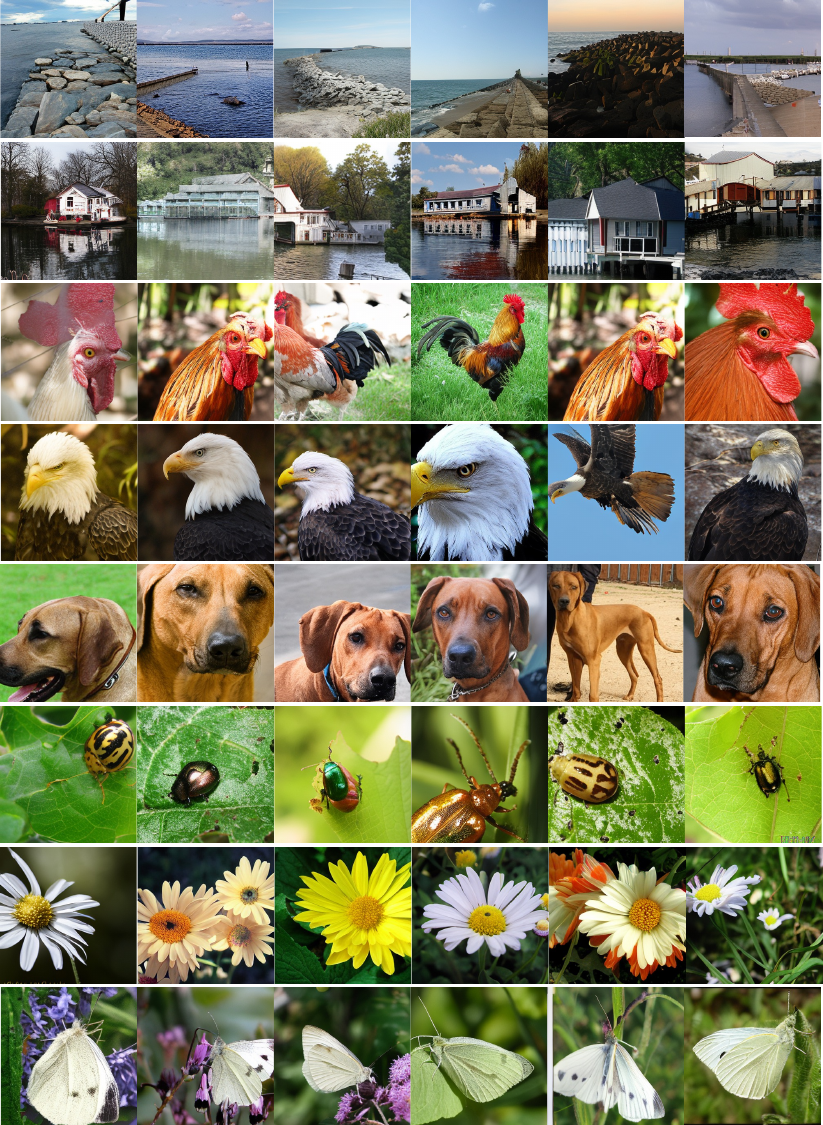}
  \vspace{-0.5pc}
\caption{Visualization of class-conditional samples generated by Hita. Best viewed with zoom-in.}
\label{fig:class-conditional-gen}
\vspace{-2pc}
\end{figure*}

In this section, we present more visualization samples including zero-shot style transfer (see Fig.~\ref{fig:zero-shot-style-transfer}), image in-painting (see Fig.~\ref{fig:in-painting}), and class-conditional image generation (see Fig.~\ref{fig:class-conditional-gen}). For optimal clarity, please zoom in.

\clearpage

\end{document}